\documentclass{article}

\usepackage{graphicx} 
\usepackage{float}
\usepackage{authblk}

\usepackage[colorlinks=true, urlcolor=blue, linkcolor=red]{hyperref}
\title{When Brain-inspired AI Meets AGI}
\newcommand*\samethanks[1][\value{footnote}]{\footnotemark[#1]}
\author[1]{Lin Zhao \thanks{Co-first authors}}
\author[2]{Lu Zhang \samethanks}
\author[1]{Zihao Wu}
\author[3]{Yuzhong Chen}
\author[1]{Haixing Dai}
\author[2]{Xiaowei Yu}
\author[1]{Zhengliang Liu}
\author[4]{Tuo Zhang}
\author[4]{Xintao Hu}
\author[3]{Xi Jiang}
\author[5]{Xiang Li}
\author[2]{Dajiang Zhu}
\author[6,7,8]{Dinggang Shen}
\author[1]{Tianming Liu \thanks{Corresponding author: tianming.liu@gmail.com}}
\affil[1]{School of Computing, The University of Georgia, Athens 30602, USA}
\affil[2]{Department of Computer Science and Engineering, The University of Texas at Arlington, Arlington 76019, USA}
\affil[3]{MOE Key Laboratory for Neuroinformation, School of Life Science and Technology, University of Electronic Science and Technology of China, Chengdu 611731, China}
\affil[4]{School of Automation, Northwestern Polytechnical University, Xi'an 710072, China}
\affil[5]{Department of Radiology, Massachusetts General Hospital and Harvard Medical School, Boston 02115, USA}
\affil[6]{School of Biomedical Engineering, ShanghaiTech University, Shanghai 201210, China}
\affil[7]{Shanghai United Imaging Intelligence Co., Ltd., Shanghai 200230, China}
\affil[8]{Shanghai Clinical Research and Trial Center, Shanghai, 201210, China}

\date{}

\begin{document}

\maketitle

\begin{abstract}
Artificial General Intelligence (AGI) has been a long-standing goal of humanity, with the aim of creating machines capable of performing any intellectual task that humans can do. To achieve this, AGI researchers draw inspiration from the human brain and seek to replicate its principles in intelligent machines. Brain-inspired artificial intelligence is a field that has emerged from this endeavor, combining insights from neuroscience, psychology, and computer science to develop more efficient and powerful AI systems. In this article, we provide a comprehensive overview of brain-inspired AI from the perspective of AGI. We begin with the current progress in brain-inspired AI and its extensive connection with AGI. We then cover the important characteristics for both human intelligence and AGI (e.g., scaling, multimodality, and reasoning). We discuss important technologies toward achieving AGI in current AI systems, such as in-context learning and prompt tuning. We also investigate the evolution of AGI systems from both algorithmic and infrastructural perspectives. Finally, we explore the limitations and future of AGI.
    
\end{abstract}

\section{Brain-inspired AI and AGI}

The human brain is widely considered one of the most intricate and advanced information-processing systems in the world. It comprises over 86 billion neurons, each capable of forming up to 10,000 synapses with other neurons, resulting in an exceptionally complex network of connections that allows for the proliferation of intelligence. Along with the physiological complexity, the human brain exhibits a wide range of characteristics that contribute to its remarkable functional capabilities. For example, it can integrate data from multiple sensory modalities, such as vision, hearing, and touch, allowing it to form a coherent perception of the world. The brain's ability to perform parallel processing is also essential for efficiently handling multiple information streams simultaneously. This is fulfilled via the connections and real-time communications among different brain regions, though the mechanism is not fully understood. Besides, the brain is highly adaptable, capable of reorganizing its structure and function in response to changing environments and experiences. This property, known as neuroplasticity, enables the brain to learn and develop new skills throughout life. The human brain is also notable for its high-level cognitive functions, such as problem-solving, decision-making, creativity, and abstract reasoning, supported by the prefrontal cortex, a brain region that is particularly well-developed in humans.

Creating artificial general intelligence (AGI) system that has human-level or even higher intelligence and is capable of performing a wide range of intellectual tasks, such as reasoning, problem-solving, and creativity, is the pursuit of humanity for centuries, which can date back to the mid-20th century. In the 1940s, pioneers such as Alan Turing developed early ideas about computing machines and the potential for them to simulate human thinking \cite{turing2009computing}. From then on, seeking to replicate the principles of human intelligence in artificial systems has significantly promoted the development of AGI and the corresponding applications. These principles include the structure and function of neural networks, the plasticity of synaptic connections, the dynamics of neural activity, and more. In 1943, McCulloch and Pitts proposed the very first mathematical model of an artificial neuron \cite{mcculloch1943logical}, also known as McCulloch-Pitts (MCP) Neuron. Inspired by the Hebbian theory of synaptic plasticity, Frank Rosenblatt came up with the perceptron, a major improvement over the MCP neuron model \cite{rosenblatt1961principles}, and showed that by relaxing some of the MCP’s rules artificial neurons could actually learn from data. However, the research of artificial neural network had stagnated until the backprogation was proposed by Werbos in 1975 \cite{werbos1974beyond}. Backpropagation was inspired by the way the brain modifies the strengths of connections between neurons to learn and improve its performance through synaptic plasticity. Backpropagation attempts to mimic this process by adjusting the weights (synaptic strengths) between neurons in an artificial neural network. Despite this early proposal, backpropagation did not gain widespread attention until the 1980s, when researchers such as David Rumelhart, Geoffrey Hinton, and Ronald Williams published papers that demonstrated the effectiveness of backpropagation for training neural networks \cite{rumelhart1985learning}.

Convolutional neural networks (CNNs) is one of the most widely used and effective types of neural networks for processing visual information \cite{lecun1995convolutional}. CNNs are also inspired by the hierarchical organization of the visual cortex in the brain, which can be traced back to the work of David Hubel and Torsten Wiesel in 1960s\cite{hubel1962receptive}. In the visual cortex, neurons are arranged in layers, with each layer processing visual information in a hierarchical manner. The input from the retina is first processed by a layer of simple cells that detect edges and orientations, and then passed on to more complex cells that recognize more complex features such as shapes and textures. Their work provided insights into how the visual system processes information and inspired the development of CNNs that could mimic this hierarchical processing process. Attention mechanisms in artificial neural network are also inspired by the way human brain selectively attend to certain aspects of sensory input or cognitive processes, allowing us to focus on important information while filtering out irrelevant details \cite{posner1990attention}. Attention has been studied in the fields of psychology and neuroscience for many years, and its application to artificial intelligence significantly advances our steps towards AGI. The ``Transformer" model, based on self-attention mechanism, has become the basis for many state-of-the-art artificial neural networks such as BERT~\cite{devlin2018bert} and GPT \cite{radford2018improving}. By adapting self-attention mechanisms into image processing, Vision Transformer (ViT) \cite{dosovitskiy2020image} model demonstrated state-of-the-art performance in various computer vision (CV) tasks by representing the image as a sequence of patches.

Recently, more and more evidence suggests that artificial neural networks (ANNs) and biological neural networks (BNNs) may share common principles in optimizing network architecture. For example, the property of small-world in brain structural and functional networks has been extensively studied in the literature \cite{bassett2006small,bullmore2009complex,bassett2017small}. In a recent study, neural networks based on Watts-Strogatz (WS) random graphs with small-world properties have demonstrated competitive performances compared to hand-designed and NAS-optimized models \cite{xie2019exploring}. Additionally, post-hoc analysis has shown that the graph structure of top-performing ANNs, such as CNNs and MLP, is similar to that of real BNNs, such as the network in the macaque cortex \cite{you2020graph}. Chen et al. proposed a unified and biologically-plausible relational graph representation of ViT models, finding that model performance was closely related to graph measures and the ViT has high similarity with real BNNs \cite{chen2022unified}. Zhao et al. synchronized the activation of ANNs and CNNs and found that CNNs with higher performance are similar to BNNs in terms of visual representation activation \cite{zhao2022coupling}. Liu et al. coupling the artificial neurons in BERT model with the biological neurons in the human brain, and found that artificial neurons can carry mearningful linguistic/semantic information and anchor to their biological neurons signatures with interpretablility in a neurolinguistic context. Zhou et al. treated each hidden dimension in Wav2Vec2.0 as a artificial neuron and connected them with biological counterparts in the human brain, suggesting a close relationship between two domain in terms of neurolinguistic information.

Following this trend, there has been growing interest in developing brain-inspired artificial intelligence by drawing inspiration from some human brain prior knowledge, such as the organization of brain structure and function. For example, Huang et al. \cite{huang2022bi} proposed a brain inspired adversarial visual attention network (BI-AVAN) which imitates the biased competition process in human visual system to decode the human visual attention. Inspired by the core-periphery organization of human brain, Yu et al. proposed a core-periphery principle guided vision transformer model (CP-ViT) for image recognition with improved performances and interpretability. Similarly, Zhao et al. implemented the core-periphery principle in the design of network wiring patterns and the sparsification of the convolution operation. The proposed core-periphery principle guided CNNs (CP-CNNs) demonstrate the effectiveness and superiority compared to CNNs and ViT-based methods. Another groups of studies opted for spiking neural networks (SNNs) \cite{ghosh2009spiking} which closely emulate the behavior of biological neurons in the brain. For example, SNNs was employed to map and understand the spatio-temploral brain data \cite{kasabov2014neucube}, decode and understand muscle activity from electroencephalography signals \cite{kumarasinghe2021brain}, and brain-machine interfaces \cite{dethier2013design,kumarasinghe2020deep}.

Brain-inspired AI has also contributed to the development of hardware architectures that mimic the structure and function of the brain. Neuromorphic computing, a field of study that aims to design computer hardware that emulates the biological neurons and synapses, has also gained increasing attention in recent years \cite{merolla2014million,benjamin2014neurogrid,zhang2016creating,davies2018loihi,roy2019towards,pei2019towards}. Neuromorphic chips are designed to process information in a parallel and distributed way, similar to the way the brain works, which can lead to significant improvements in efficiency and speed compared to traditional computing architectures. Some of the neuromorphic chips, such as IBM's TrueNorth chip \cite{akopyan2015truenorth} and Intel's Loihi chip \cite{davies2018loihi}, use spiking neural networks to process information in a way that is closer to how the brain processes information. These chips have been used for a wide range of applications, including image and speech recognition \cite{indiveri2000neuromorphic}, robotics \cite{sandamirskaya2022neuromorphic}, and autonomous vehicles \cite{viale2022lanesnns}. The advancement of brain-inspired hardware also provides a potential for significant breakthroughs in the field of AGI by paving the road for generalized hardware platforms \cite{pei2019towards}.

\begin{figure}
\begin{center}
\includegraphics[width=1.0\textwidth]{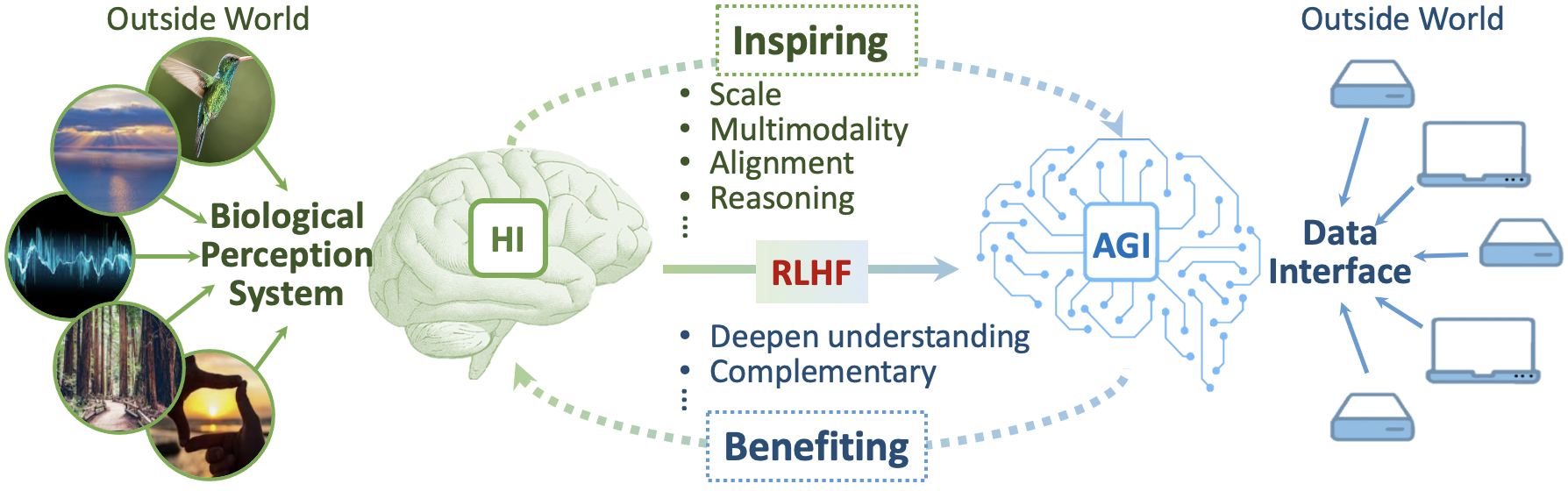}
\end{center}
\caption{The development of artificial general intelligence (AGI) has been greatly inspired by the study of human intelligence (HI). In turn, AGI has the potential to benefit human intelligence. Current language models such as ChatGPT and GPT-4 use reinforcement learning with human feedback (RLHF) to align their behavior with human values. As we continue to study and understand both human intelligence and AGI, these two systems will become increasingly intertwined, enhancing and supporting each other in new and exciting ways.} 
\label{main}
\end{figure}

Overall, brain-inspired AI plays a crucial role in the development of AGI (Figure \ref{main}). By drawing inspiration from the human brain, researchers can create algorithms and architectures that are better suited to handle complex, real-world problems that require a high degree of flexibility and adaptability. This is especially important for AGI, which aims to develop machines that can perform a wide range of tasks, learn from experience, and generalize their knowledge to new situations. The human brain is one of the most complex information-processing system known to us, and it has evolved over millions of years to be highly efficient and effective in handling complex tasks. By studying the brain and developing AI systems that mimic its architecture and function, researchers can create AGI that is more sophisticated and adaptable, bringing us closer to the ultimate goal of creating machines that can match or surpass human intelligence. In turn, AGI also has the potential to benefit human intelligence and deepen our understanding of intelligence. As we continue to study and understand both human intelligence and AGI, these two systems will become increasingly intertwined, enhancing and supporting each other in new and exciting ways.

\section{Characteristics of AGI}

\subsection{Scale}

The scale of brains varies greatly across different animal species, ranging from a few thousand neurons in the case of simple invertebrates such as nematode worms, to over 86 billion neurons in the case of humans. For example, the brain of a fruit fly contains around 100,000 neurons, and the brain of a mouse contains around 70 million neurons. For primates, macaque monkey's brain has around 1.3 billion neurons \cite{christensen2007neocortical} while the chimpanzee's brain have around 6.2 billion neurons \cite{dicke2016neuronal}. Compared to other animals, the human brain is the most complex and sophisticated biological structure known to science, containing over 86 billion neurons. The scale of the brain, i.e., the number of neurons, is often correlated with the cognitive abilities of the animal and considered as a factor of intelligence \cite{dicke2016neuronal,herculano2012remarkable}. The size and complexity of the brain regions associated with specific cognitive functions, such as language or memory, are often directly related to the number of neurons they contain \cite{huttenlocher1979synaptic, rakic1995small,sporns2011human}.

The relationship between the number of neurons and cognitive abilities is also relevant for large language models (LLMs) such as GPT-2 and GPT-3. While GPT-2 has 1.5 billion parameters and was trained on 40 gigabytes of text data, GPT-3 has 175 billion parameters and was trained on 570 gigabytes of text data. This significant increase in the number of parameters has enabled GPT-3 to outperform GPT-2 on a range of language tasks, demonstrating an increase in its ability to perform complex language tasks. In fact, GPT-3 has been shown to achieve human-like performance on several natural language processing benchmarks, such as question-answering, language translation, and text completion tasks. Its size and capacity for natural language processing has made it a powerful tool for various applications, including chatbots, content generation, and language translation.

This trend is similar to the way larger brains are associated with more complex cognitive functions in animals. As LLMs continue to scale up, it is expected that they will become even more capable of few-shot learning of new skills from a small number of training examples, similar to how animals with larger brains have more sophisticated cognitive abilities. This correlation suggests that scale may be a crucial factor in achieving artificial general intelligence. However, it's worth noting that the number of parameters alone does not determine the intelligence of an LLM. The quality of the training data, the training process, and the architecture of the model also play important roles in its performance. As researchers continue to improve LLMs and explore new ways to advance AGI, it will be interesting to see how the relationship between the number of parameters and cognitive abilities evolves.

\begin{figure}[H]
\begin{center}
\includegraphics[width=1.0\textwidth]{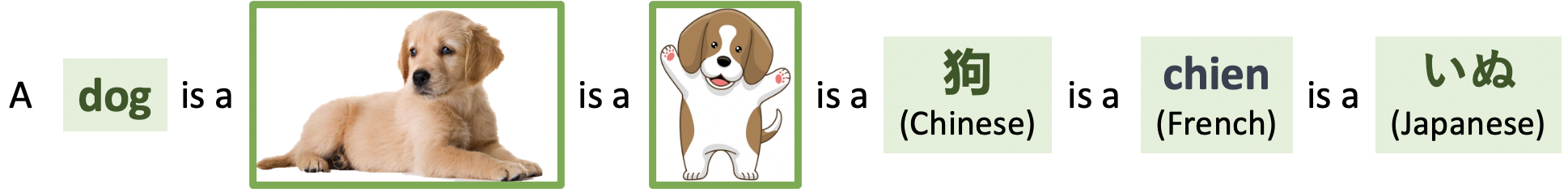}
\end{center}
\caption{The concept ``dog" represented in different modalities.} 
\label{concept_dog}
\end{figure}
\subsection{Multimodality}
The human brain's ability to process and integrate information from multiple sensory modalities simultaneously is a remarkable feat. This feature allows individuals to understand the world around them through various sources of information, such as sight, sound, touch, taste, and smell. Moreover, processing multimodal information enables people to make more accurate and comprehensive assessments of their environment and communicate effectively with others. Consequently, successful learning from multiple modalities can enhance human cognitive abilities. As we strive to create advanced artificial general intelligence (AGI) systems that surpass human intelligence, it is crucial that they are capable of acquiring and ingesting knowledge from various sources and modalities to solve tasks that involve any modality. For instance, an AGI should be able to utilize knowledge learned from images and the knowledge base to answer natural language questions, as well as use knowledge learned from text to perform visual tasks. Ultimately, all modalities intersect through universal concepts, such as the concept that a dog is a dog, regardless of how it is represented in different modalities (Figure \ref{concept_dog}).

To build multimodal AI systems, a promising approach is to incorporate training signals from multiple modalities into LLMs. This requires aligning the internal representations across different modalities, enabling the AI system to integrate knowledge seamlessly. For instance, when an AI system receives an image and related text, it must associate the same object or concept between the modalities. Suppose the AI sees a picture of a car with text referring to its wheels. In that case, the AI needs to attend to the part of the image with the car wheels when processing the text mentioning them. The AI must ``understand" that the image of the car wheels and the text referring to them describe the same object across different modalities. 

In recent years, multimodal AI systems have been experimenting with aligning text/NLP and vision into an embedding space to facilitate multimodal decision-making. Cross-modal alignment is essential for various tasks, including text-to-image and image-to-text generation, visual question answering, and video-language modeling. In the following section, we provide a brief overview of these prevalent workloads and the corresponding state-of-the-art models.

\subsubsection{Text-to-image and image-to-text generation}
CLIP \cite{radford2021learning}, DALL-E \cite{ramesh2021zero},  and their successor GLIDE \cite{nichol2021glide}, VisualGPT\cite{Chen_2022_CVPR} and Diffusion\cite{rombach2022high} are some of the most well-known models that tackle image descriptions (image-to-text generation) and text-to-image generation tasks. CLIP is a pre-training method that trains separate image and text encoders and learns to predict which images in a dataset are associated with various descriptions. Notably, similar to the ``Halle Berry" neuron in humans \cite{quiroga2005invariant}, CLIP has been found to have multimodal neurons that activate when exposed to both the classifier label text and the corresponding image \cite{goh2021multimodal}, indicating a fused multimodal representation. DALL-E, on the other hand, is a variant of GPT-3 with 13 billion parameters that takes text as input and generates a sequence of images to match the input text. The generated images are then ranked using CLIP. GLIDE, an evolution of DALL-E, still uses CLIP to rank the generated images, but the image generation is accomplished using a diffusion model \cite{rombach2022high}. Stable Diffusion is also based on diffusion models while it operates on the latent space of powerful pre-trained autoencoders and thus in limited computational resources while maintaining their quality and flexibility. The VisualGPT is the evolution of GPT-2 from a single language model to a multimodal model with a self-resurrecting activation unit to produce sparse activations that prevent accidental overwriting of linguistic knowledge.

\subsubsection{Visual question answering}
Visual question answering is a critical application of multimodal learning that requires a model to correctly respond to a text-based question based on an image. The VQA dataset \cite{antol2015vqa} presents this task, and teams at Microsoft Research have developed some of the leading approaches for it. One of these approaches is METER \cite{dou2022empirical}, a general framework for training performant end-to-end vision-language transformers using a variety of sub-architectures for the vision encoder, text encoder, multimodal fusion, and decoder modules. This flexibility allows METER to achieve state-of-the-art performance across a range of tasks. Another promising approach is the Unified Vision-Language pretrained Model (VLMo) \cite{bao2022vlmo}, which uses a modular transformer network to jointly learn a dual encoder and a fusion encoder. Each block in the network contains a pool of modality-specific experts and a shared self-attention layer, offering significant flexibility for fine-tuning. This architecture has shown impressive results on several benchmark datasets.

\subsubsection{Video-language modeling}
Traditionally, AI systems have struggled with video-based tasks due to the high computational resources required. However, this is beginning to change, thanks to efforts in the domain of video-language modeling and other video-related multimodal tasks, such as Microsoft's Project \href{https://www.microsoft.com/en-us/research/project/project-florence-vl/}{Florence-VL}. In mid-2021, Project Florence-VL introduced ClipBERT \cite{lei2021less}, a combination of a CNN and a transformer model that operates on sparsely sampled frames. It is optimized in an end-to-end fashion to solve popular video-language tasks. Subsequent evolutions of ClipBERT, such as VIOLET \cite{fu2021violet} and SwinBERT\cite{lin2022swinbert}, have introduced Masked Visual-token Modeling and Sparse Attention to improve the state-of-the-art in video question answering, video retrieval, and video captioning. While each of these models has unique features, they all utilize a transformer-based architecture. Typically, this architecture is coupled with parallel learning modules to extract data from various modalities and unify them into a single multimodal representation.

Recently, the emergence of GPT-4 has taken multimodal research to a new level. According to the latest official research paper \cite{bubeck2023sparks}, GPT-4 not only exhibits high proficiency in various domains, including literature, medicine, law, mathematics, physical sciences, and programming but also fluently combines skills and concepts from multiple domains, demonstrating impressive comprehension of complex ideas. Furthermore, GPT-4's performance in all of these tasks is remarkably close to human-level performance and often surpasses prior models such as ChatGPT. Given the breadth and depth of GPT-4's capabilities, it could be viewed as an early version (albeit incomplete) of an artificial general intelligence (AGI) system. 

It is important to note that, in contrast to single-modality LLMs, multimodal LLMs exhibit superior performance not only in cross-modal tasks but also in single-modality tasks. For instance, the integration of multimodality in GPT-4 results in better performance in textual tasks compared to ChatGPT \cite{bubeck2023sparks}. This aligns with the way humans perceive the world through multiple sensory modalities.

\subsection{Alignment}
While some LLMs like BERT \cite{devlin2018bert}, GPT \cite{radford2018improving}, GPT-2\cite{radford2019language}, GPT-3 \cite{brown2020language}, and Text-to-Text Transfer Transformer (T5) \cite{raffel2020exploring} have achieved remarkable success in specific tasks, they still fall short of true Artificial General Intelligence (AGI) due to their tendency to exhibit unintended behaviors. For example, they might generate biased or toxic text, make up facts, or fail to follow user instructions. The main reason behind these issues is the misalignment between the language modeling objective used for many recent LLMs and the objective of safely and helpfully following user instructions. Therefore, while these models have made significant advancements, they are not yet capable of emulating human-like reasoning, decision-making, and understanding. To achieve AGI, it's crucial to align language models with the user's intention. This alignment will enable LLMs to function safely and helpfully, making them more reliable for complex tasks that require nuanced decision-making and understanding. To achieve this, there is a need for better algorithms that steer agents towards human values while fostering cross-disciplinary collaborations to clarify what ``human values" mean.

Recent developments in large language models (LLMs), such as Sparrow \cite{glaese2022improving}, InstructGPT \cite{ouyang2022training}, ChatGPT, and GPT-4, have addressed the issue of alignment with human instructions using reinforcement learning from human feedback (RLHF). Reinforcement learning is a type of machine learning where the model learns to make decisions based on feedback it receives in the form of rewards. The goal of the model is to maximize its total reward over time. RLHF uses human preferences as a reward signal to fine-tune the LLMs and enable LLMs to learn and improve from human feedback, which tries to predict what answers the humans will react positively to and helps in reducing unintended behaviors and increasing their reliability for complex tasks. Since the model learns from humans in real-time, it becomes better and better at predicting. At the end of the training process, AI systems start to imitate humans. RLHF has shown promising results and is a significant step towards developing LLMs that can function safely and helpfully, aligning with human values and intentions.

\subsection{Reasoning}
Reasoning plays a crucial role in human intelligence and is essential for decision-making, problem-solving, and critical thinking. Previous study \cite{herculano2009human} have explored factors that influence intelligence levels by comparing different attributes of brains across various mammalian species. The findings suggest that cognitive abilities are primarily centered on the absolute number of neurons. Among mammals, the human brain has the highest number of neurons, which gives it superior reasoning and intelligence abilities compared to other species. Recently, a similar phenomenon has also emerged in the LLMs. It has been observed that the LLMs exhibit emergent behaviors, such as the ability to reason, when they reach a certain size \cite{wei2022emergent}. To enhance LLM's reasoning abilities, two major types of approaches have been developed. The first type, known as Prompt-based methods \cite{kojima2022large, wei2022chain,wang2022self,li2022advance}, is more widely researched and involves leveraging appropriate prompts to better stimulate the reasoning abilities that LLMs already possess. The second type of approaches involves introducing program code into the pre-training process \cite{zhang2022automatic,suzgun2022challenging,li2022advance}, where it is trained alongside text to further enhance the LLM's reasoning ability. The two approaches have fundamentally different directions: using code to enhance LLM reasoning abilities represents a strategy of directly enhancing LLM reasoning abilities by increasing the diversity of training data; while the Prompt-based approach does not promote LLM's own reasoning abilities, but rather provides a technical method for LLM to better demonstrate this ability during problem-solving.

Currently, most existing works in the field of large language models (LLM) reasoning adopt prompt-based methods, which can roughly be divided into three technical routes. The first approach is the zero-shot Chain of Thought (CoT), proposed by \cite{kojima2022large}. This method is simple and effective, involving two stages. In the first stage, a prompt phrase, \textit{``Let's think step by step"}, is added to the question, and the LLM outputs a specific reasoning process. In the second stage, the reasoning process output by the LLM in the first stage is concatenated with the question, and the prompt phrase, \textit{``Therefore, the answer (arabic numerals) is"}, is added to obtain the answer. Such a simple operation can significantly increase the effectiveness of the LLM in various reasoning tasks. For example, Zero-shot-CoT achieves score gains from 10.4\% to 40.7\% on arithmetic benchmark GSM8K \cite{kojima2022large}. The second approach is the Few-Shot CoT \cite{wei2022chain}, which is currently the main direction of LLM reasoning research. The main idea of Few-Shot CoT is straightforward: to teach the LLM model to learn reasoning, provide some manually written reasoning examples, and clearly explain the specific reasoning steps one by one before obtaining the final answer in the examples. These manually written detailed reasoning processes are referred to as Chain of Thought Prompting. The concept of CoT was first explicitly proposed by \cite{wei2022chain}. Although the method is simple, the reasoning ability of the LLM model has been greatly improved after applying CoT. The accuracy of the GSM8K mathematical reasoning dataset increased to around 60.1\% \cite{wei2022chain}. Based on CoT, subsequent works \cite{wang2022self,li2022advance} have expanded from a single Prompt question to multiple Prompt questions, checked the correctness of intermediate reasoning steps, and improved the accuracy of multiple outputs using weighted voting. These improvements have continuously raised the accuracy of the GSM8K test set to around 83\%. The third 
 approach is ``Least-to-most prompting" \cite{zhou2022least}. The core idea is to decompose a complex reasoning problem into several easier-to-solve subproblems that can be solved sequentially, whereby solving a given subproblem is facilitated by the answers to previously solved subproblems. After solving each subproblem, we can derive the answer to the original problem from the answers to the subproblems. This idea is highly consistent with the divide-and-conquer algorithm that humans use to solve complex problems. As our understanding of the brain and LLMs continues to deepen, it will be interesting to investigate whether these two network systems share an optimal structure.

\section{Important Technology}
 Language models, such as LLMs, rely on several crucial techniques include zero-shot prompting, few-shot prompting, in-context learning, and instruct. The underlying expectation of these techniques is that AI systems can learn new tasks rapidly by leveraging what they have learned in the past, much like humans do. Through the use of these techniques, language models can be trained to perform a wide range of tasks, from generating coherent text to answering complex questions, with greater accuracy and efficiency. Ultimately, these advancements bring us closer to realizing the potential of AI to assist and augment human intelligence in new and exciting ways. Of these techniques, instruct serves as the interface utilized by ChatGPT, where users provide task descriptions in natural language, such as ``Translate this sentence from Chinese to English." Interestingly, zero-shot prompting was initially the term used for Instruct. During the early stages of zero-shot prompting, users faced difficulty expressing tasks clearly, leading them to try various wordings and sentences repeatedly to achieve optimal phrasing. Presently, Instruct involves providing a command statement to facilitate LLM understanding. In Context Learning and few-shot prompting share similar purposes, which involve presenting LLMs with a few examples as templates to solve new problems. Accordingly, this article places emphasis on introducing In Context Learning and Instruct.
\subsection{In-Context Learning}
The foremost capability of the human brain resides in its robust learning capacity, enabling the execution of cognitive, computational, expressive, and motor functions predicated on linguistic or visual prompts, often with minimal or no examples. This attribute is central to the attainment of human-level Artificial General Intelligence (AGI). Recent advancements in large-scale AGI models, specifically GPT-4, have demonstrated such a promising capability. They are pretrained on massive multimodal datasets, capturing a wide range of tasks and knowledge while understanding diverse prompts from both linguistic and visual domains. This enables in context learning akin to the human brain's working mode, and driving AGI into real-world applications. In fact, following the emergence of large-scale models like GPT-4 and Midjourney V5, many industries, such as text processing and illustration, have witnessed disruptive scenarios where AGI liberates human labor. These models leverage prior knowledge acquired from pretraining across various tasks and context, allowing rapid adaptation to novel tasks without the need for extensive labeled data for fine-tuning, which is a critical challenge in fields like medical\cite{liu2023deid} and robotics\cite{liu2023digital} where labeled data is often limited or even unavailable.

In the context of AGI, in-context learning denotes the model's capacity to comprehend and execute new tasks by providing a limited number of input-output pair examples\cite{dai2023chataug} within prompts or merely a task description. Prompts facilitate the model's apprehension of the task's structure and patterns, while in context learning exhibit similarities to explicit fine-tuning at the prediction, representation, and attention behavior levels. This allows them to generalize to and perform new tasks even better without further training or fine-tuning\cite{dai2022can} and reduces the likelihood of overfitting downstream labeled training data.

Despite the absence of fine-tuning requirements in these large-scale AGI models, the trade-offs include increased computational costs due to their massive parameter scale and the potential need for expert knowledge in formulating effective prompts with examples during inference. Potential solutions entail hardware advancements and the integration of more refined domain-specific knowledge during the pretraining phase.

\subsection{Prompt and Instruction Tuning}

Like human infants generally acquire various concepts about the world mostly by observation, with very little direct intervention\cite{lecun2022path}, the large-scale AGI models also gain wide-ranging knowledge after initial large-scale unsupervised training and have achieved remarkable generalization performance. The prompt and instruction tuning-based methods allow the pretrained models to achieve zero-shot learning in numerous downstream applications\cite{sanh2021multitask}.

The human brain is always an efficient and orderly processor, providing targeted feedback for the current task rather than speaking nonsense. In addition to the brain's innate pursuit of efficiency, moral and legal constraints ingrained in human development also ensure that human interactions are orderly and beneficial. For AGI models to reach human-level performance, producing truthful and harmless results based on instructions is an essential requirement. Although current large-scale AGI models have powerful generative capabilities, a key question is whether these capabilities can be aligned with users' intent. This is important as it relates to whether the model can produce satisfactory results for users, even in situations where tasks and prompts are unseen and unclear. Additionally, as these models become more widely used, untruthful and toxic outputs must be effectively controlled.

InstructGPT\cite{ouyang2022training} is at the forefront in this regard.  In order to improve the quality of model outputs, supervised training is conducted using human-provided prompts and demonstrations. The outputs generated by different models are then collected and ranked by humans based on their quality. The models are further fine-tuned using a technique known as Reinforcement Learning from Human Feedback (RLHF)\cite{christiano2017deep}, which utilizes human preferences as rewards to guide the learning process. In addition, To avoid InstructGPT aligning exclusively with human tasks at the expense of neglecting classical NLP tasks, a small amount of the original data used to train GPT-3 (InstructGPT's foundation) is mixed in. Recent research\cite{chung2022scaling, wang2022benchmarking} has demonstrated that incorporating larger-scale and more diverse task instruction datasets can further enhance model performance.

\section{Evolution of AGI}
Artificial General Intelligence (AGI) refers to an advanced level of artificial intelligence (AI) that mirrors human-like abilities in understanding, learning, and applying knowledge across a broad spectrum of tasks and subject areas. Unlike \textit{narrow AI} (e.g., a tailored convolutional neural network for face recognition), which is designed to perform specific tasks, AGI is capable of adapting to new situations, transferring domain knowledge, and exhibiting human-like cognitive abilities beyond streamlined and formatted task-solving workflows in the current literature ~\cite{goertzel2014artificialm,hodson2019deepmind}. Overall, AGI could demonstrate remarkable versatility and adaptability, 

While the scientific community has not yet accomplished genuine AGI, the advancement made in artificial intelligence and its subfields (e.g., deep learning), has laid the foundation for further exploration and the quest towards achieving AGI.. Here's a brief overview of the history of AGI:

\subsection{Early Days of AI}
The AGI concept can be traced back to the work of Alan Turing, who proposed the idea that machines could think and learn like humans in a 1950 manuscript \textit{``Computing Machinery and Intelligence"} ~\cite{turing2009computing}. Turing's ideas laid the groundwork for AI development and computer science in general.

In 1956, The Dartmouth workshop ~\cite{kline2010cybernetics}, organized by pioneers such as John McCarthy, Marvin Minsky, Nathaniel Rochester, and Claude Shannon, marked the inception of AI as an academic discipline. Their objective was to develop machines that could imitate human intelligence. This collective endeavor played a significant role in shaping the future course of the AI community.

The initial optimism and enthusiasm in the field led to the development of early AI programs such as the General Problem Solver ~\cite{nilsson2009quest}, , Logic Theorist ~\cite{gugerty2006newell}, and ELIZA ~\cite{weizenbaum1966eliza}. However, these AI systems were limited in scope and unpractical for large scale real-world applications. A period known as the ``AI winter" occurred because of a decline in funding and interest in artificial intelligence research. This was due to the lack of significant progress made in the field and the unrealistic claims made by some researchers. Reduced funding support, in turn, led to further decline in progress and a decrease in the number of published research papers. 

The renewed interest in AI was brought about by artificial neural networks that were modeled after the structure and function of the human brain ~\cite{shanmuganathan2016artificial,lippmann1987introduction}. The backpropagation algorithm, introduced by Rumelhart, Hinton, and Williams in 1986 ~\cite{rumelhart1986learning}, allowed neural networks to learn more efficiently and laid down a solid foundation for modern neural networks.

In addition, the emergence of machine learning methods such as  support vector machines ~\cite{cortes1995support}, decision trees ~\cite{quinlan1986induction}, and ensemble methods ~\cite{opitz1999popular} proved to be powerful tools for pattern recognition and classification. These methods propelled AI research and empowered practical applications, further driving the field forward.. 

\subsection{Deep Learning and Modern AGI}
The development of deep learning, enabled by revolutionary advancements in computing power and the availability of large datasets, has led to notable advancements in the field of AI. Breakthroughs in computer vision, natural language processing, and reinforcement learning are bringing the prospect of AGI closer to becoming a tangible reality. In particular, the Transformer architecture ~\cite{vaswani2017attention}, introduced by Vaswani et al. in 2017, revolutionized language modeling by leveraging self-attention mechanisms to capture global dependencies and contextual relationships between words in a sequence. This breakthrough laid the foundation for the rise of pre-trained language models, such as GPT-3 ~\cite{brown2020language}, and vision transformer (ViT) based models ~\cite{dosovitskiy2020image} in computer vision. This shared architectural ancestry has also paved the way for the  the development of transformer-based multimodal models ~\cite{liu2022survey,hu2021unit} .

Since 2019, the introduction of large-scale language models like GPT-2 ~\cite{radford2019language} and GPT-3 ~\cite{brown2020language}, both based on the transformer architecture, have demonstrated impressive natural language understanding and generation capabilities. While these models are not yet AGI, they represent a significant step of progress towards achieving this goal. Both GPT-2 and GPT-3 are based on GPT ~\cite{radford2018improving}, a decoder-only pre-trained language model that leverages self-attention mechanisms to capture long-range dependencies between words in a sequence.

Recent advancements in AI give rise to groundbreaking extensions of the GPT models, such as ChatGPT and GPT-4. ChatGPT builds upon the success of GPT-3, incorporating reinforcement learning from human feedback (RLHF) to generate outputs that properly align with human values and preferences. The chatbot interface of CharGPT has enabled millions of users to engage with AI in a more natural way, and it has been applied in diverse use cases such as essay writing, question answering, search, translation~\cite{qin2023chatgpt}, data augmentation~\cite{dai2023chataug}, computer-aided diagnosis ~\cite{wang2023chatcad} and data de-identification ~\cite{liu2023deid}. On the other hand, GPT-4 represents a significant leap forward in the GPT series, with a massive set of 10 trillion parameters. It is capable of advanced math, logic reasoning. In addition, the model excels in standard examinations such as the USMLE, LSAT, and GRE ~\cite{openai2023gpt4}. GPT-4 has broad applicability and is expected to solve an unprecedented range of problems. Its development is a testament to the tremendous progress made in the pursuit of AGI.

\subsection{The Infrastructure of AGI}
One key aspect of AGI is the infrastructure required to support it. Neural networks have been a major component of this infrastructure, and their development has evolved significantly since their inception in the 1940s and 1950s. Early artificial neural networks (ANNs) were limited in their capabilities due to their simple linear models. However, the backpropagation algorithm \cite{werbos1974beyond}, created by Werbos in 1975, revolutionized the field by making it possible to efficiently train neural networks with multiple layers, including the perceptron. This algorithm calculates gradients, which are used to update the weights of the neural network during training, allowing it to learn and improve its performance over time. Since the development of backpropagation, neural network research has advanced rapidly, with the creation of more sophisticated architectures and optimization algorithms. Today, neural networks are used for a wide range of tasks, including image classification, natural language processing, and prediction, and continue to be an active area of research in machine learning and artificial intelligence.

In addition to algorithm, the progress in hardware, particularly the development of graphics processing units (GPUs) and tensor processing units (TPUs), has made it possible to train deep neural networks efficiently, leading to the widespread adoption of deep learning. This progress has enabled the development of more powerful neural networks, which can tackle increasingly complex problems and has accelerated the research and development of AGI. For example, Microsoft's investment of \$1 billion in OpenAI in 2019 enabled the creation of a dedicated Azure AI supercomputer, one of the world's most powerful AI systems. This supercomputer is equipped with over 285,000 CPU cores and over 10,000 GPUs, and it is designed to support large-scale distributed training of deep neural networks. Such investments in infrastructure are critical for the development of AGI.

Recent advancements in AI models, particularly the GPT series \cite{radford2018improving, radford2019language}, have provided valuable insights into the infrastructure requirements for AGI development. To train AI models, three essential components of AGI infrastructure are required: massive data requirements, computational resources, and distributed computing systems. GPT models, including GPT-2 and GPT-3, were primarily trained on large-scale web datasets, such as the WebText dataset, which consisted of 45 terabytes of text data before preprocessing and deduplication, reduced to around 40 gigabytes of text after preprocessing. Training a GPT model requires powerful hardware and parallel processing techniques, as exemplified by GPT-3, which was trained using large-scale distributed training across multiple GPUs, consuming a significant amount of computational resources and energy. Developing an AGI model, such as GPT-4, necessitates distributed computing techniques. While the specific distributed computing systems used for training GPT models may not be publicly disclosed, TensorFlow, PyTorch, and Horovod are distributed computing frameworks that facilitate the implementation of these techniques. Researchers and developers can use these frameworks to distribute the training process across multiple devices, manage device communication and synchronization, and efficiently utilize available computational resources.

\section{Discussion}

\subsection{Limitations}
While significant progress has been made in the development of AGI and brain-inspired AI, there are still several limitations that need to be overcome before we can achieve true human-level intelligence in machines. Some of these limitations include:

\textbf{Limited understanding of the human brain}: Despite significant advancements in neuroscience and brain-inspired AI, we still have a limited understanding of how the human brain works. This makes it challenging to create machines that can fully replicate human intelligence.

\textbf{Data efficiency:} Current AGI and brain-inspired AI systems require vast amounts of training data to achieve comparable performance to humans. This is in contrast to humans, who can learn from relatively few examples and generalize to new situations with ease. How to efficiently learn from few samples is still an opening question.

\textbf{Ethics}: There are also ethical considerations to consider with AGI. As these systems become more intelligent, they may be able to make decisions that have far-reaching consequences. Ensuring that these decisions are aligned with human values and ethical principles is critical for preventing unintended harm.

\textbf{Safety}: Safety is also a significant concern with AGI. Ensuring that these systems do not cause unintended harm, either through malicious intent or unintentional mistakes, is critical for their widespread adoption. Developing robust safety mechanisms and ensuring that AGI systems are aligned with human values is essential.

\textbf{Computational Cost}: Current LLM models requires massive computational resources to train and operate, making it challenging to develop and deploy in a wide range of scenarios. Meanwhile, the computational cost can limit the number of researchers and organizations working in the field, which may slow the progress towards AGI. Additionally, the energy consumption of AGI systems can be prohibitively high, making them unsustainable from an environmental perspective.

\subsection{Future of AGI}
The future of AGI is an exciting and rapidly evolving field. While the development of AGI remains a challenge, it has the potential to revolutionize many aspects of our lives, from healthcare to transportation to education. One potential avenue for advancing AGI is through the creation of more powerful and sophisticated AGI foundation models. Recent breakthroughs in natural language processing, computer vision, knowledge graph, and reinforcement learning have led to the development of increasingly advanced AGI models such as ChatGPT and GPT-4. These models have shown impressive capabilities in various applications. Further advances in AGI foundation model research, as well as improvements in hardware and computational algorithms, are very likely to accelerate the development of AGI.

Another approach to developing AGI is through the integration of different AI systems and technologies across multiple domains, including adding human in the loop through reinforcement learning from expert feedback. For example, combining natural language processing with computer vision and robotics under the guidance of human experts could lead to the creation of more versatile and adaptable intelligent systems. This integration could also help overcome the limitations of current AI systems, which are often specialized in specific domains and lack the flexibility to transfer knowledge across domains.

The development of AGI also requires the development of novel approaches for machine learning, such as more efficient instruct methods, in-context learning algorithms, and reasoning paradigm, particularly by learning from the human brain via brain-inspired AI. These approaches aim to enable machines to learn from unstructured data without the need of labeling them and rapidly generalize from a few examples, which is crucial for enabling machines to learn and adapt to new tasks and environments.

Finally, ethical and societal implications of AGI development must be considered, including issues related to bias, privacy, and security. As AGI becomes more powerful and pervasive, it is essential to ensure that it is developed and used in a responsible and ethical manner that benefits society as a whole and aligns well with human value. Overall, while the development of AGI remains a challenge, it has the potential to revolutionize many aspects of our lives and bring significant benefits to society and humanity. Ongoing research and development in AGI will continue to drive progress towards the ultimate goal of creating truly intelligent machines.

\section{Conclusion}

In this article, we provided a comprehensive overview of brain-inspired AI from the perspective of AGI, covering its current progress, important characteristics, and technological advancements towards achieving AGI. We also discussed the evolution, limitations and the future of AGI. In conclusion, brain-inspired AI is a promising field that has the potential to unlock the mysteries of human intelligence and pave the way for AGI. While significant progress has been made in recent years, there is still much work to be done to achieve AGI. It will require advances in technology, algorithms, and hardware, as well as continued collaboration across multiple disciplines. Nonetheless, the pursuit of AGI is an important and worthwhile endeavor that has the potential to transform our world in unprecedented ways. We hope this survey provides a valuable contribution to this exciting field and inspires further research and development toward the ultimate goal of AGI.

\bibliographystyle{vancouver}
\bibliography{mybib} 
\end{document}